\documentclass[reprint,3p]{elsarticle}

\usepackage{subcaption}
\usepackage{color}
\usepackage[version=3]{mhchem}
\usepackage[detect-all]{siunitx}
\DeclareSIUnit{\Molar}{\textsc{m}}
\usepackage{amsmath,amssymb}

\usepackage{lineno,hyperref}
\begin{document}

\begin{frontmatter}

\title{Belousov-Zhabotinsky liquid marbles in robot control}

\author[label1]{Michail-Antisthenis Tsompanas\corref{cor1}}
\address[label1]{Unconventional Computing Centre, University of the West of England, Bristol BS16 1QY, UK\fnref{label1}}
\cortext[cor1]{Corresponding author}
\ead{antisthenis.tsompanas@uwe.ac.uk}

\author[label1]{Claire Fullarton}
\author[label1]{Andrew Adamatzky}

\begin{abstract}
We show how to control the movement of a wheeled robot using on-board liquid marbles made of Belousov-Zhabotinsky solution droplets coated with polyethylene powder. Two stainless steel, iridium coated electrodes were inserted in a marble and the electrical potential recorded was used to control the robot's motor. We stimulated the marble with a laser beam. It responded to the stimulation by pronounced changes in the electrical potential output. The electrical output was detected by robot. The robot was changing its trajectory in response to the stimulation. The results open new horizons for applications for oscillatory chemical reactions in robotics.
\end{abstract}

%\begin{keyword}
%Microbial fuel cells, cellular non-linear network, spatial models
%\end{keyword}

\end{frontmatter}

\section{Introduction}
\label{sec1}

%\textcolor{cyan}{Reaction-diffusion chemical processes have attracted vast amounts of interest within the scientific community over the past few decades~\cite{epstein1996nonlinear,sagues2003nonlinear,mikhailov2006control}. Belousov-Zhabotinsky (BZ) type reactions are examples of these excitable oscillatory non-linear media. The BZ reaction, if left unstirred, can exhibit interesting spatio-temporal patterns as a result of chemical excitation wave-fronts e.g. target waves, spiral waves and sometimes localised wave-fragments. These oxidation waves have been used for image processing and computation. Wet electronics and computing circuits prototyped in BZ media include chemical diodes~\cite{DBLP:journals/ijuc/IgarashiG11}, Boolean gates~\cite{steinbock1996chemical, sielewiesiuk2001logical}, neuromorphic architectures~\cite{ gorecki2006information, gentili2012belousov, takigawa2011dendritic, stovold2012simulating,  gruenert2015understanding} and associative memory~\cite{stovold2016reaction,stovold2017associative}, wave-based counters~\cite{gorecki2003chemical}, arithmetic circuits~\cite{costello2011towards, sun2013multi, zhang2012towards, suncrossover, digitalcomparator}.}

Belousov-Zhabotinsky (BZ) reaction is an oscillatory chemical reaction, a model of excitable non-linear medium~\cite{epstein1996nonlinear,sagues2003nonlinear,mikhailov2006control}. A non-stirred BZ solution exhibits interesting spatio-temporal patterns as a result of chemical excitation wave-fronts, e.g. target waves, spiral waves and localised wave-fragments. The oxidation waves have been used for image processing and computation. Wet electronics and computing circuits prototyped in BZ medium include chemical diodes~\cite{DBLP:journals/ijuc/IgarashiG11}, Boolean gates~\cite{steinbock1996chemical, sielewiesiuk2001logical}, neuromorphic architectures~\cite{ gorecki2006information, gentili2012belousov, takigawa2011dendritic, stovold2012simulating,  gruenert2015understanding} and associative memory~\cite{stovold2016reaction,stovold2017associative}, wave-based counters~\cite{gorecki2003chemical},  arithmetic circuits~\cite{costello2011towards, sun2013multi, zhang2012towards, suncrossover, digitalcomparator}. 

%\textcolor{cyan}{Using the BZ reaction as a media for controlling robots has been theoretically studied in models of excitable automata lattices supplied with propulsive cilia~\cite{adamatzky2002phototaxis}. Recently, an experimental prototype has been produced which uses a chemical processor to navigate a robot around obstacles in an arena~\cite{adamatzky2003experimental}. However, the processor in this case required manual images of the whole experimental arena to be acquired.}

BZ controller for robots have been studied theoretically in the models of excitable automata lattices supplied with propulsive \textcolor{black}{cilia}~\cite{adamatzky2002phototaxis}. A chemical processor to navigate a robot around obstacles in an arena has been prototyped in~\cite{adamatzky2003experimental}. This processor, however, required images of the whole experimental arena to be prepared by a human operation in an off-line mode. The first real time BZ controller for a robot was designed and prototyped in~\cite{adamatzky2004experimental}. In this case, a thin layer of BZ medium contained within a Petri dish was mounted onto a wheeled robot. Direction towards a source of stimulation was inferred, via an optical interface, from the 2D patterns of oxidation wave-fronts. Another example a BZ robotic controller is the closed loop control of a robotic hand with a thin layer BZ reactor~\cite{yokoi2004excitable}. The closed loop is achieved with photo-sensors placed underneath the Petri dish where the excitation of the BZ medium occurs from the movement of the robotic fingers. The way the robotic fingers react, in turn, is controlled by a micro-controller receiving data from the photo-sensors. The developed hybrid system was able to deliver highly complex behaviour by using just three sensors and three of the five fingers. Recently, the use of BZ gels was proposed to assemble millimetre-sized soft robots that exhibit photo-taxis, while not using any other kind of device to move around \cite{dayal2014directing}. With the simulation of the chemical, along with the mechanical motion of the gels, their capabilities were unveiled. These worm-like gels can follow complicated routes based on different intensities of light, perform periodic movement resembling cilia and self-organise in groups.

%Initial proposals of using non-linear chemical media, known for their capacity of unconventional processing of data \cite{adamatzky2001computing}, such as a navigator (a tool for controlling the trajectory) of a mobile device via decentralised propulsive actuators, were based on simulation results \cite{adamatzky2002phototaxis}. The simulations were involving a cellular automata lattice that the outer cells were acting as photosensitive sensors and all the other cells were contributing in a combined movement of the entity. The inner cells' states were updated based on the states in each cells' vicinity. All possible local rules were studied (256 rules for 8 neighbours and two possible states) and the behaviours of the navigators were categorised in four groups based on their success. 

%A chemical reaction--diffusion processor was realised, in order to be utilised as a path planner for a robot \cite{adamatzky2003experimental}. The goal of this unconventional processor was to identify the route that would be further away from obstacles, thus alleviating the event of a possible collision. The motivation behind using this type of processor was under the vision of enabling future generation, non-silicon robots, namely soft robotics. These processors would have to be aware of the whole experimental arena that the robot would test and the chemicals would have to be prepared by a human operator in an off-line procedure.

Previously, the BZ medium has been utilised as an isolated system that needs specialised interfaces and data processing tools. We propose a hybrid system where the chemical system provides information to conventional electronics that control the movement of the robotic system through a direct electrical connection. Thus, this is a further step towards the final goal of an autonomous next generation of soft robots. 

Previous prototypes of BZ controllers employed BZ in a Petri dish, which posed difficulties with manipulation and portability of the prototypes~\cite{adamatzky2004experimental,yokoi2004excitable}. Therefore, to overcome these difficulties, we decided to encapsulate ferroin-catalysed BZ solution in a liquid marble. A liquid marble (LM) is microlitre liquid \textcolor{black}{droplets} encapsulated in a hydrophobic powder coating~\cite{Aussillous2001}. This approach enables us to transfer and manipulate the BZ LM controller without wetting the underlying surface. Our scoping studies showed that the BZ media encapsulated in LMs exhibits `classical' chemical excitation wave patterns, with mainly trigger waves observed~\cite{fullarton2018belousov}.
The BZ media has been reported to be sensitive to illumination. Toth et al~\cite{toth2000wave} experimentally demonstrated that  visible light of the appropriate frequency, in their case it was a He–Ne laser with wavelength \SI{632.8}{nm}, initiates oxidation in the ferroin-catalysed BZ reaction. 
\textcolor{black}{Moreover, visible light of different wavelengths is proved to initiate or inhibit the dynamics of ferroin- or ruthenium-catalyzed BZ medium due to the photochemical properties of the catalyst \cite{ Steinbock1993,Vanag2000,Kadar1997,Gaspar1983,Hanazaki1995}.}
We also use a laser beam to stimulate BZ LMs onboard of a Zumo robot. 

The BZ LMs were mounted on and electrically interfaced with the Zumo robot~\cite{zumo}. \textcolor{black}{The alternation in the dynamics of the reagents inside the BZ LM can be monitored potentiometrically with two iridium coated stainless steel electrode. Several paradigms of studies that use electrical potential to monitor the oscillation in a BZ system were previously published \cite{Crowley1986, Gaspar1983, Petrov1993}.} The robot is attractive in its simplicity of design and control. It has been used previously in studying route-following by klinokinesis, inspired by the navigation skills of desert ants~\cite{ants2015}, randomised algorithm mimicking biased lone exploration in roaches ~\cite{hart2015low}, and the self optimisation procedure on a line-tracing application by using a evolutionary computing algorithm~\cite{zahn2016optimization}.

\section{Methods}
\label{methods}

\begin{figure}[!tb]
    \centering
%   \subfigure[]{\includegraphics[width=0.3\textwidth]{BZ_Marble_Electrodes}\label{marble}}
%   \subfigure[]{\includegraphics[width=0.3\textwidth]{BZMarblesRobot}\label{robot}}
%\subfigure[]{\includegraphics[width=0.2\textwidth]{MarblePierced}\label{marblepiereced}}
%   \subfigure[]{\includegraphics[width=0.475\textwidth]{Robot}\label{snap01}}
%   \subfigure[]{\includegraphics[width=0.45\textwidth]{RobotLaser}\label{snap02}}
\includegraphics[width=0.75\textwidth]{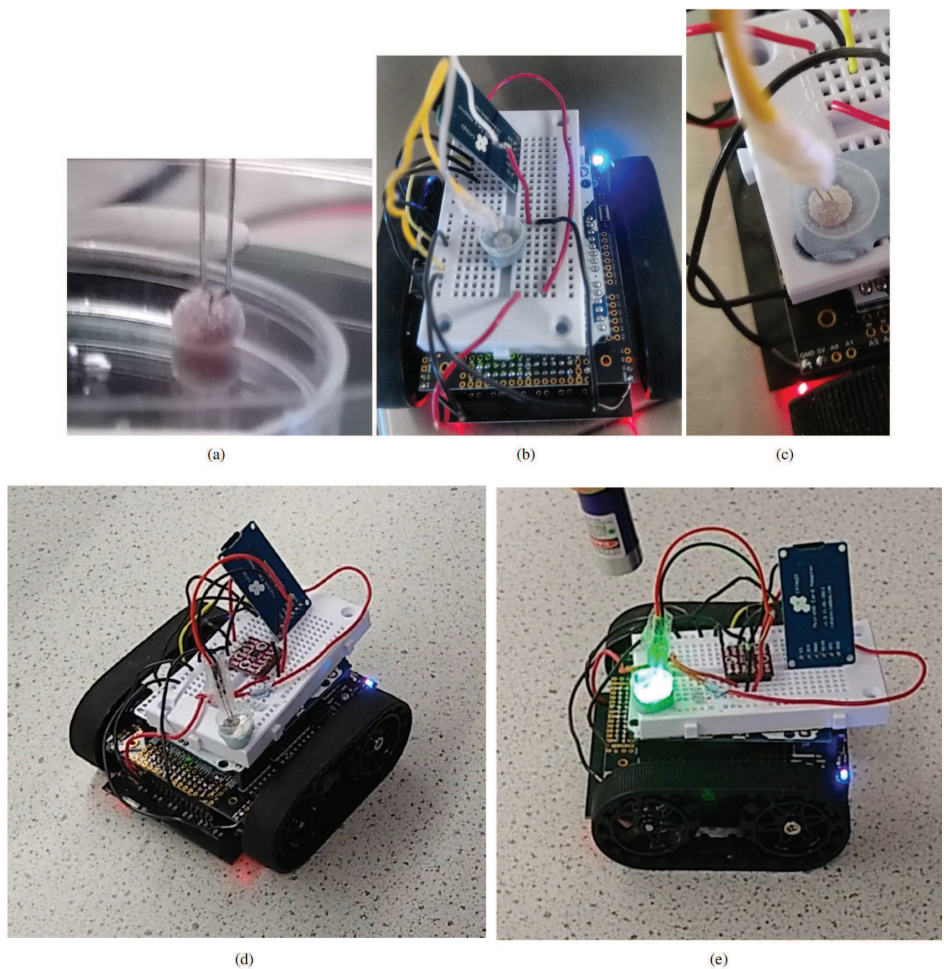}
    \caption{\textbf{Experimental setup.} 
    (a)~Electrical connection to record the potential of the BZ marble. 
    (b)~BZ marble rests on the robot platform.
    (c) Close up shot of two BZ LM on-board the robot platform.
    (d) Robot is controlled by BZ marble. 
    (e) BZ LM is stimulated with laser beam. 
    }
    \label{snap}
  \end{figure}

Belousov-Zhabotinsky (BZ) liquid marbles (LMs) were \textcolor{black}{produced by coating droplets of BZ solution with ultra high density polyethylene (PE) powder (Sigma Aldrich, CAS 9002-88-4, Product Code 1002018483, particle size $100 \mu m$). The BZ solution was} prepared using the method \textcolor{black}{reported by Field~\cite{Field1979},} omitting the surfactant Triton X. \SI{18}{\Molar} Sulphuric acid \ce{H2SO4} (Fischer Scientific), sodium bromate \ce{NaBrO3}, malonic acid \ce{CH2(COOH)2}, sodium bromide \ce{NaBr} and \SI{0.025}{\Molar} ferroin indicator \textcolor{black}{(Sigma Aldrich)} were used as received. Sulphuric acid (\SI{2}{\ml}) was added to deionised water (\SI{67}{\ml}), to produce \SI{0.5}{\Molar} \ce{H2SO4}, \ce{NaBrO3} was added to the acid to yield \SI{70}{\ml} of stock solution.

Stock solutions of \SI{1}{\Molar} malonic acid and \SI{1}{\Molar} NaBr were prepared by dissolving \SI{1}{\gram} in \SI{10}{\ml} of deionised water. In a \SI{50}{\ml} beaker, \SI{0.5}{\ml} of \SI{1}{\Molar} malonic acid was added to \SI{3}{\ml} of the acidic \ce{NaBrO3} solution. \SI{0.25}{\ml} of \SI{1}{\Molar} NaBr was then \textcolor{black}{added} to the beaker, which produced bromine. \textcolor{black}{The solution was set aside} until it was clear and colourless (ca.\ \SI{3}{\minute}) before adding \textcolor{black}{\SI{0.5}{\ml} of \SI{0.025}{\Molar} ferroin indicator.}

%Stock solutions of \SI{1}{\Molar} malonic acid and \SI{1}{\Molar} NaBr were prepared by dissolving \SI{1}{\gram} in \SI{10}{\ml} of deionised water. In a \SI{50}{\ml} beaker, \SI{0.5}{\ml} of \SI{1}{\Molar} malonic acid was added to \SI{3}{\ml} of the acidic \ce{NaBrO3} solution. \SI{0.25}{\ml} of \SI{1}{\Molar} NaBr was then \textcolor{black}{added} to the beaker, which produced bromine. We set aside the solution until it was clear and colourless (this took ca.\ \SI{3}{\minute}), before adding the ferroin indicator. \SI{0.5}{\ml} of \SI{0.025}{\Molar} ferroin was added to the beaker. 

%As coating, we used ultra high density polyethylene (PE) (Sigma Aldrich, CAS 9002-88-4, Product Code 1002018483, particle size $100 \mu m$). To prepare a BZ LM, a $75 \mu L$ droplet was pipetted, from a height of ca.\ \SI{2}{\mm}, onto a powder bed of PE. The BZ droplet was rolled on the powder bed for ca.\ \SI{10}{\second} until the droplet was fully coated with powder.

\textcolor{black}{BZ LMs were prepared by pipetting a $75 \mu L$ droplet of BZ solution, from a height of ca.\ \SI{2}{\mm} onto a powder bed of PE, using a method reported previously \cite{fullarton2018belousov}}. The BZ droplet was rolled on the powder bed for ca.\ \SI{10}{\second} until it was fully coated with powder.

\textcolor{black}{For the initial experiments, which aimed to establish} the electrical potential outputs of a BZ LM, a LM was placed in Petri dish and pierced with two iridium coated stainless steel electrodes (Fig.~\ref{snap}a). \textcolor{black}{For experiments investigating} the electrical potential of a BZ LM stimulated with a laser, sub-dermal needle electrodes with twisted cables were used (\copyright~SPES MEDICA SRL Via Buccari 21 16153 Genova, Italy). Electrical \textcolor{black}{potential} outputs were recorded with an ADC-24 high resolution data logger (Pico Technology, St Neots, Cambridgeshire, UK), sampling every \SI{10}{\ms}.

%For the initially executed experiments, aiming to investigate the electrical potential outputs of a BZ LM, the marble was placed in a Petri dish and pierced with two stainless steel, iridium coated electrodes (Fig.~\ref{marble}). When studied a precise response of a BZ LM to stimulation with laser we recorded the electrical potential using sub-dermal needle electrodes with twisted cable\footnote{\copyright~SPES MEDICA SRL Via Buccari 21 16153 Genova, Italy}. Electrical activity of BZ LM for stationary experiments was recorded with ADC-24 High Resolution Data Logger\footnote{Pico Technology, St Neots, Cambridgeshire, UK}. We recorded the electrical activity one sample per \SI{10}{\ms}. 

 %We used a Zumo robot~\cite{zumo}, which is an off-the-shelf solution. The robot is developed as an Arduino shield to provide a convenient interface with its controller. The algorithm that governs a trajectory of the robot is loaded on the Arduino board and the electronics necessary to power the motors are accommodated on the robot shield (Fig.~\ref{snap01}). Light stimulation was performed using green laser pointer, wavelength \SI{532}{\nm}, for ca.\ \SI{10}{\second} (Fig.~\ref{snap02}).

\textcolor{black}{BZ LMs were mounted on the robot by rolling the LMs into plastic holders, which were subsequently attached to} the robot (Fig.~\ref{snap}b) and then the LMs pierced with two iridium coated stainless steel electrodes (Fig.~\ref{snap}c).

\textcolor{black}{The robot used was} a Zumo robot~\cite{zumo}, which was an off-the-shelf solution. The robot is developed as an Arduino shield to provide a convenient interface with its controller. The algorithm that governs the trajectory of the robot is loaded on the Arduino board and the electronics necessary to power the motors are accommodated on the robot shield (Fig.~\ref{snap}d). Light stimulation was performed using a green laser pointer, wavelength 532$nm$, 5$mW$, for ca.\ 10$s$ (Fig.~\ref{snap}e). \textcolor{black}{As previously reported ~\cite{toth2000wave}, the reduced form of the catalyst in a ferroin-catalyzed BZ medium, shows an absorption peak at 510$nm$. As a result, the choice of a wavelength of 532$nm$ is reasonably close to the peak to have significant impact in the dynamics of the reagents. A human operator have illuminated the BZ LM with a laser pen from a distance of approximately 20~$cm$. Using a FLIR ETS320 thermal camera with 0.06$^o$C resolution we found that the illumination does not lead to a substantial increase in temperature in the marble (even illumination for over 30 sec causes just 0.2$^o$C increase). }

For the on-board recording an analogue-to-digital converter was used (ADS1118 Texas Instruments Incorporated). This was because the Arduino could read only positive values of an electrical potential and its resolution was limited to \SI{4.9}{\mV}. As a result, negative values can be recorded and a higher resolution \textcolor{black}{(down to \SI{0.2}{\mV}) was} achieved. The on-board recordings were saved on to an SD-card attached to the Arduino and started \SI{3}{\second} after the activation of the robot due to initialisation procedures. \textcolor{black}{The robot is programmed with a simple algorithm to manipulate its moves in a constant way. However, this is not limiting its capabilities. Just for illustration reasons in the experiments executed in this study the algorithms dictates the robot to move 1.2$cm$ forward and turn to either direction at an angle of 3 degrees.}

\section{Results}

\begin{figure}[!tbp]
\centering
  \includegraphics[width=0.75\textwidth]{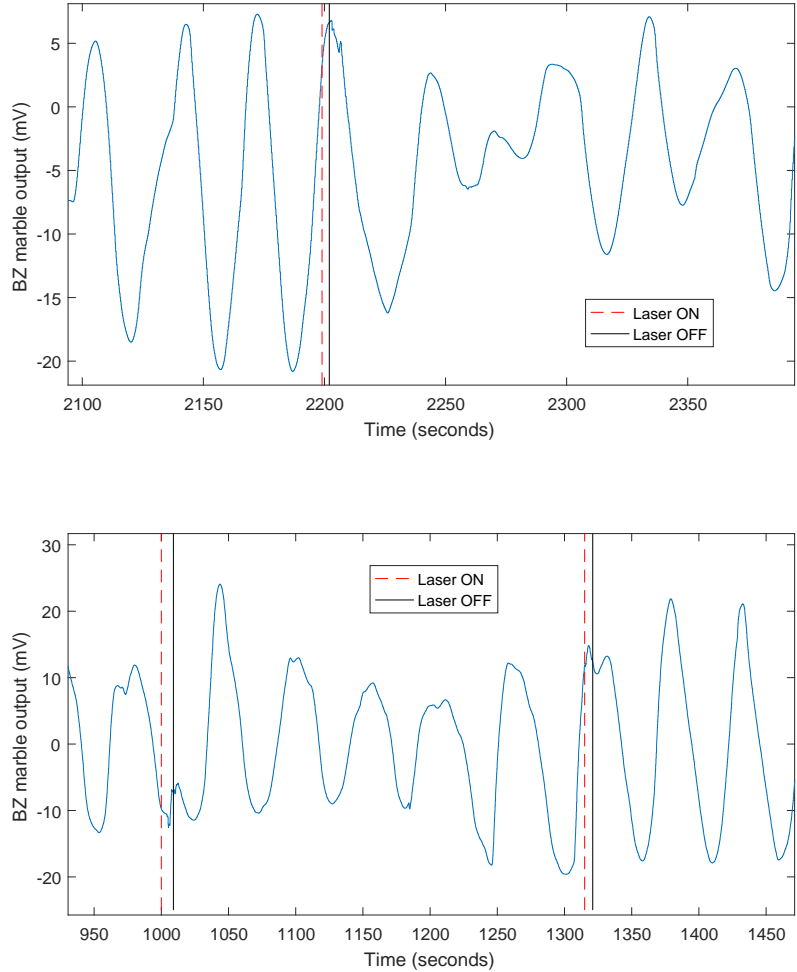}
  \label{fig:exp}
    \caption{\textbf{Potential of BZ LMs.} Dynamics of electrical potential record from BZ LMs via Pico ADC-24. Moments when the LM was illuminated with a laser are shown by vertical lines.}
    \label{fig:exp}
\end{figure}

\textcolor{black}{As the oxidation wave-fronts are travelling within the BZ LM, an electrical potential that oscillates is observed in the electrodes. The dynamics of the wave-fronts and, thus, the oscillating potential are changing in response to the LM being illuminated by a laser. More specifically, one case studied was when the LM had a potential that oscillated around a negative value and was exposed to a laser beam while at the higher point of the oscillation in the positive region (Fig.~\ref{fig:exp}(a)). The respond was inhibition of the oscillating output and a decrease of the oscillations' amplitude as realised in Fig. ~\ref{fig:exp}(a). Another case was a sudden drop of potential with no significant changing in the oscillation characteristics (Fig.~\ref{fig:exp}(b)).} %Consequently, the predominant effect is assumed to be an instantaneous and brief reduction of the output potential.  

Given the aforementioned observations of the effect the laser beam causes, we developed the algorithm that would navigate the robot by taking values of the potential from the BZ LM as follows. The algorithm, loaded to the Arduino board connected to the Zumo robot, reads the outputs from a BZ LM and if the value is positive then the robot turns left. Whereas, if the value read is negative the robot turns right. In order to avoid movement when the potential output of the BZ LM is too low, a condition of the absolute value being higher than 1mV was introduced. 
%a significant reduction in the mean voltage is detected, then the robot turns (anti)clockwise direction, depending on which one of the two BZ LMs was stimulated. 
The electrical potential of the BZ LMs is read every 2 seconds and logged on an on-board SD card for further investigation. 
%By significant reduction we define the decrease of the mean of 10 of the most current readings by a margin of 20\% compared with the mean of a fixed window of 10 readings (commencing every 5 seconds($=$ 10 readings) from the beginning of each experiment).

To enhance the comprehension of the results drawn from the robot experiments, the following figures are encoded as described here. The asterisks represent a positive potential value of the BZ LM and, hence, a left turn of the robot. Respectively, the squares in the graphs represent a negative potential value read and, hence, a right turn of the robot. The circles represent a lower value than the minimum threshold that does not dictate any movement by the robot. The dashed vertical lines represent the time slots when the laser beam stimulating the BZ LM was on, and the solid vertical lines when the laser beam was off. The $x$-axis is the time in seconds and the $y$-axis is the voltage amplitude of the BZ LM in volts.

\begin{figure}[!tbp]
    \centering
  \includegraphics[width=0.7\textwidth]{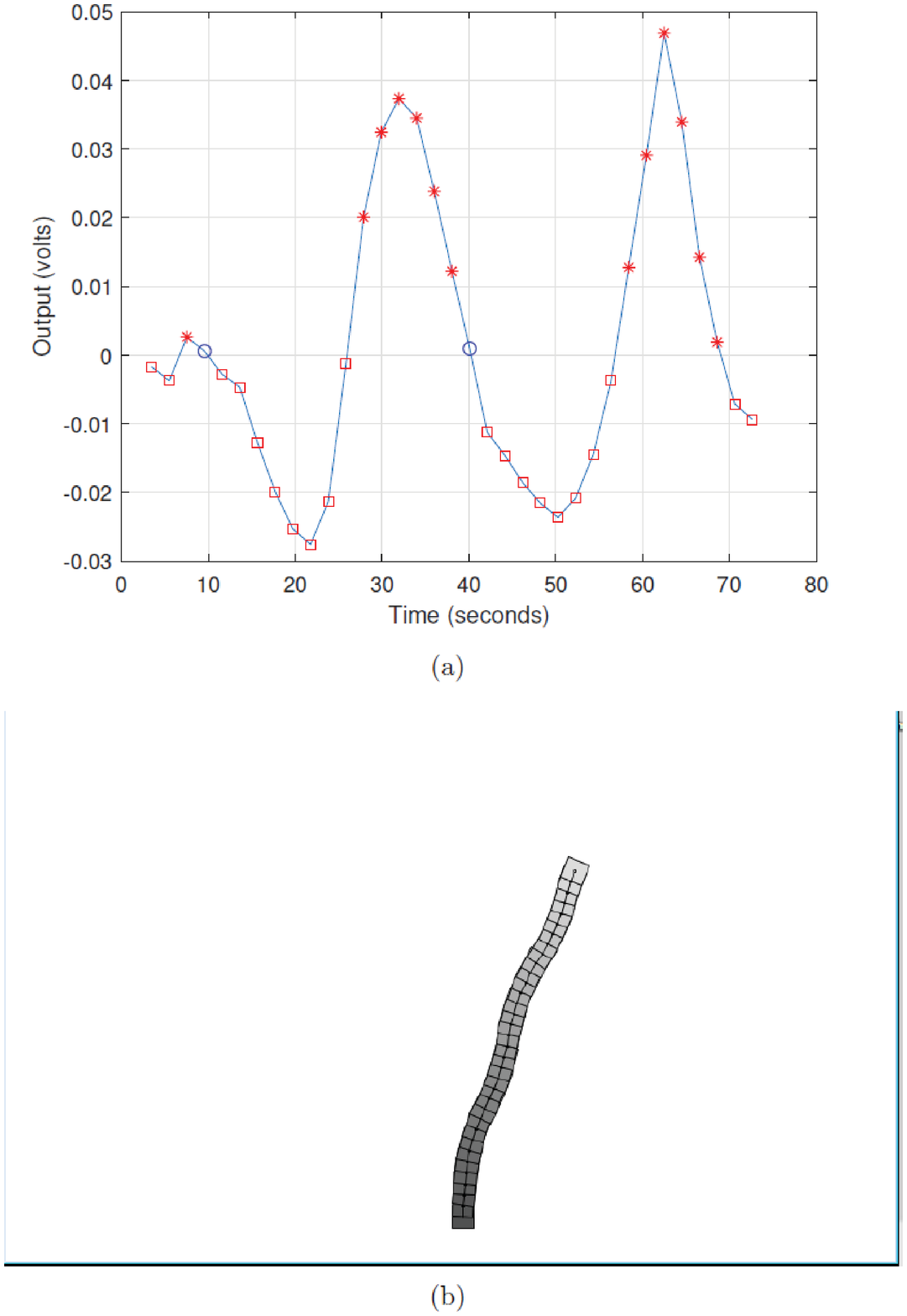}
    \caption{Results of the first experiment (with no stimulation using the laser beam). (a) Voltage output of the BZ LM and (b) trajectory of the robot. Supplementary Video BZRobot19.mp4 at \cite{BZRobotVideoZenodo}.}
    \label{fig:aexp}
\end{figure}

For the first experiment involving the robot, there was no stimulation with the laser beam. The potential output and the movement of the robot is depicted in Fig. \ref{fig:aexp}. The potential output oscillates around zero. Thus, the robot moves either towards the right direction or towards the left direction. Given that sampling points are equally distributed between negative and positive values, the robot is moving roughly towards a given direction.

\begin{figure}[!tbp]
    \centering
  \includegraphics[width=0.8\textwidth]{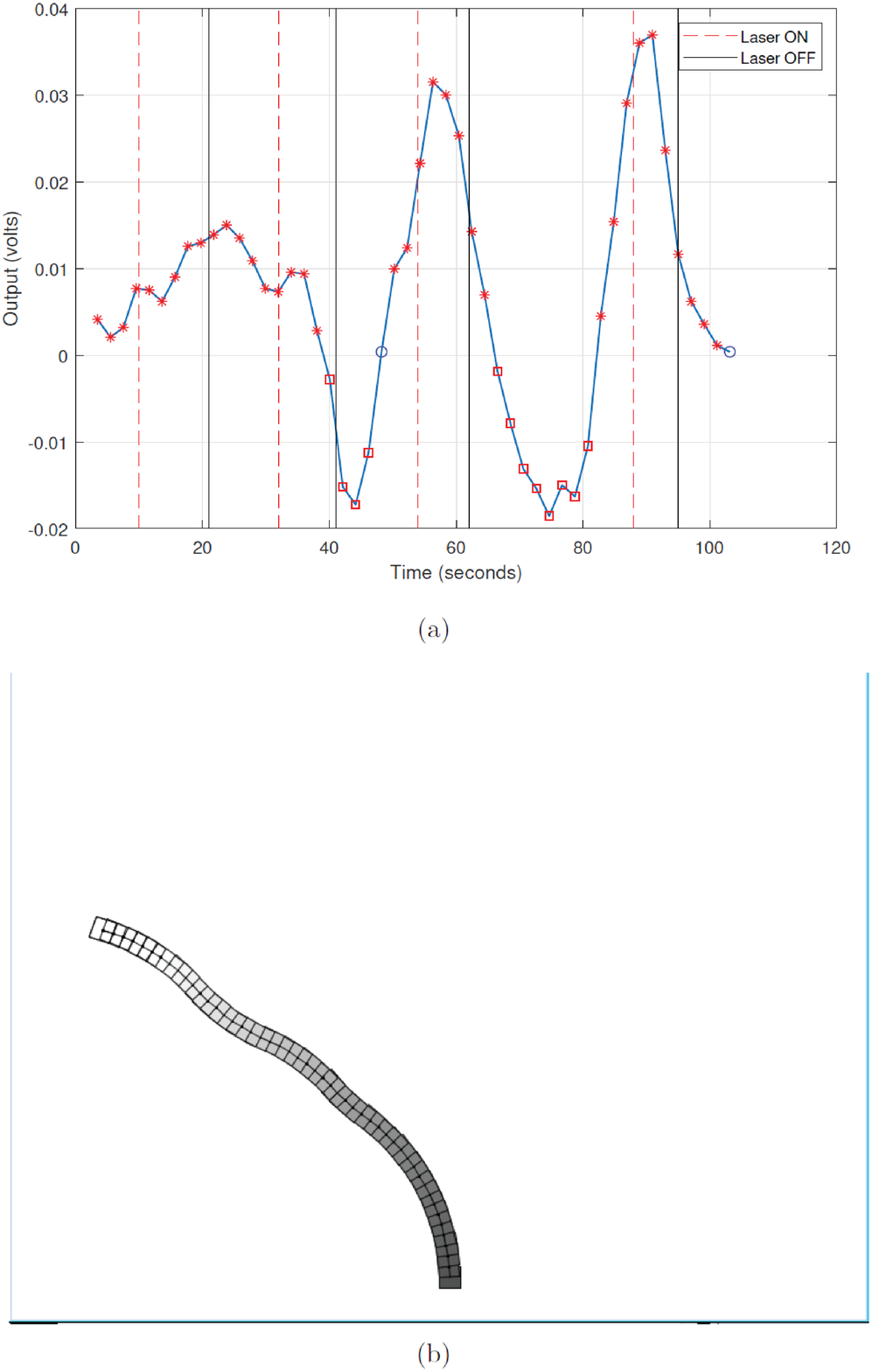}
    \caption{Results of the second experiment (with stimulation with the laser beam). (a) Voltage output of the BZ LM and (b) trajectory of the robot. Supplementary Video BZRobot20.mp4 at \cite{BZRobotVideoZenodo}.}
    \label{fig:a1exp}
\end{figure}

The second experiment with the robot was executed with the interaction of the BZ LM with a laser beam. As illustrated in the results from that experiment (Fig.~\ref{fig:a1exp}) the effects of the laser beam are altering the normal oscillation (as depicted in Fig.~\ref{fig:aexp}) of the BZ LM. The first point of stimulation (at the 10th second) hinders the oscillation and maintains the potential values in the positive area. As a result the robot keeps moving towards a left direction. The second moment of stimulation (at the 32nd second) reactivates the oscillation around zero and, thus, forcing the robot to swing its way towards a generally straight direction. However, the two remaining stimulations with the laser does not seem to have a detectable effect on the output potential of the BZ LM.

\begin{figure}[!tbp]
    \centering
    \includegraphics[width=0.8\textwidth]{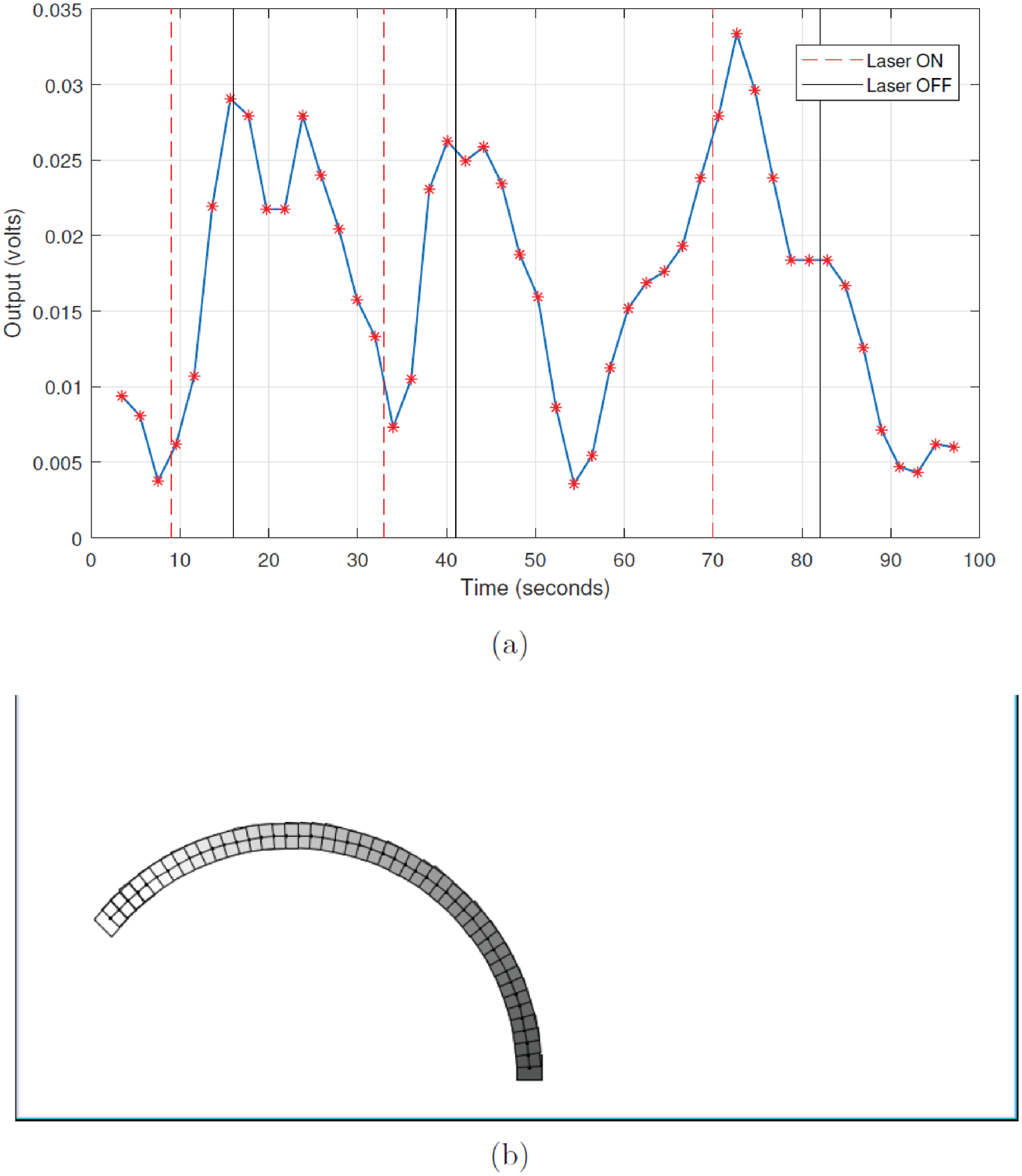}
    \caption{Results of the third experiment (with stimulation with the laser beam). (a) Voltage output of the BZ LM and (b) trajectory of the robot. Supplementary Video BZRobot21.mp4 at \cite{BZRobotVideoZenodo}.}
    \label{fig:a2exp}
\end{figure}

The results of the third experiment are featured in Fig.~\ref{fig:a2exp}. Despite the fact that all the incidents of stimulation with the laser beam have a clear effect on the oscillation and the short term amplitude of the potential, the robot moves only by turning left. The robot actually is working its way around a circle (anticlockwise), due to the fact that the potential of the BZ LM was not allowed to reach negative values, possible due to repeated initiation of oxidation wave-fronts by laser illumination. 

\begin{figure}[!tbp]
    \centering
   \includegraphics[width=0.8\textwidth]{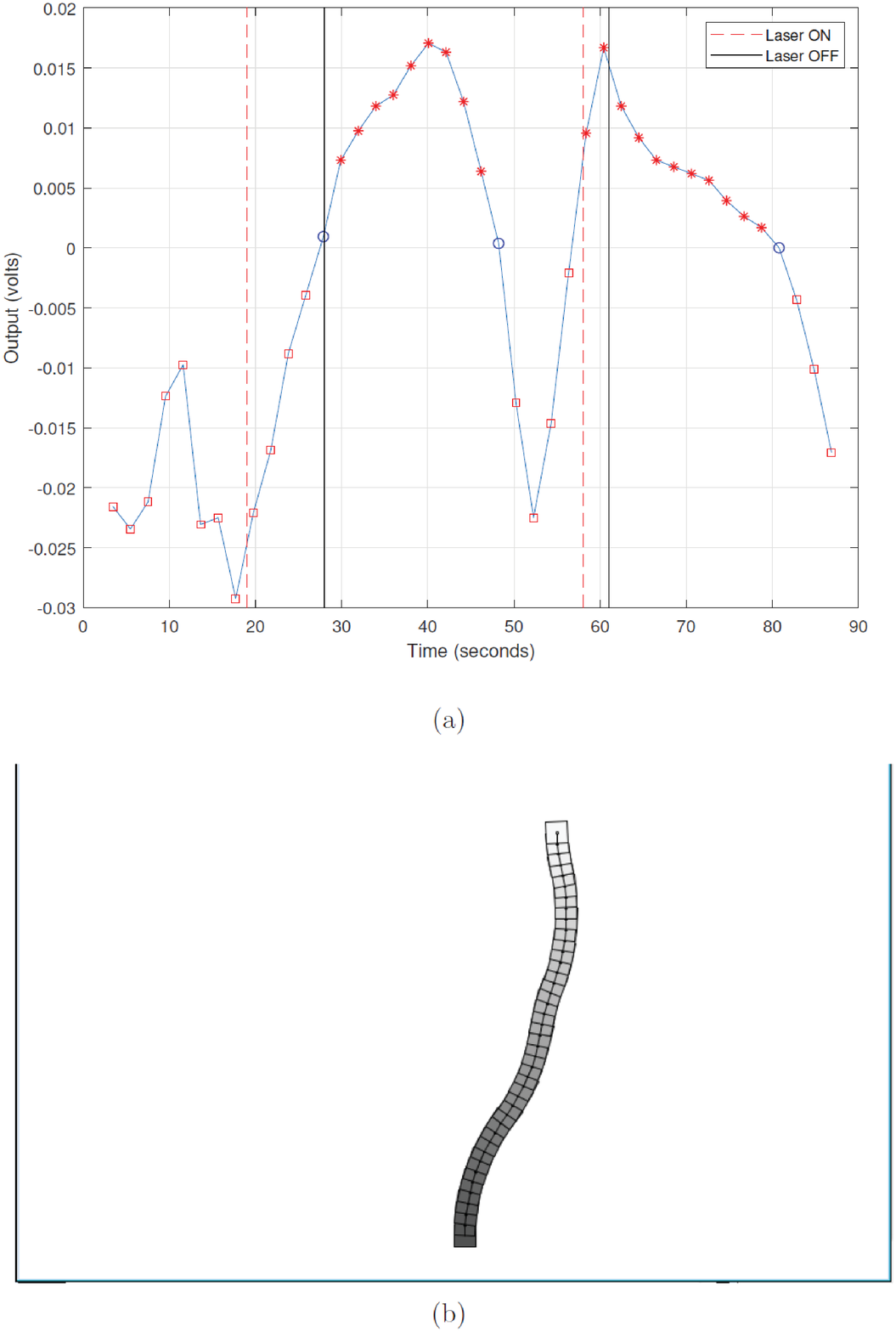}
    \caption{Results of the fourth experiment (with stimulation with the laser beam). (a) Voltage output of the BZ LM and (b) trajectory of the robot. Supplementary Video BZRobot23.mp4 at \cite{BZRobotVideoZenodo}.}
    \label{fig:a4exp}
\end{figure}

For the final experiment the results are depicted in Fig.~\ref{fig:a4exp}. Here, the output was initially oscillating within negative values. After the first stimulation with the laser beam, the potential output is constantly increasing and reaches positive values. As a result the robot stops moving on a clockwise direction and starts an anticlockwise turn. The oscillation is now around zero. However the second stimulation hinders the oscillation, with values of electrical potential remaining positive longer and, thus, the robot moves on an anticlockwise turn once more.   

\begin{figure}[!tbp]
    \centering
    \includegraphics[width=0.8\textwidth]{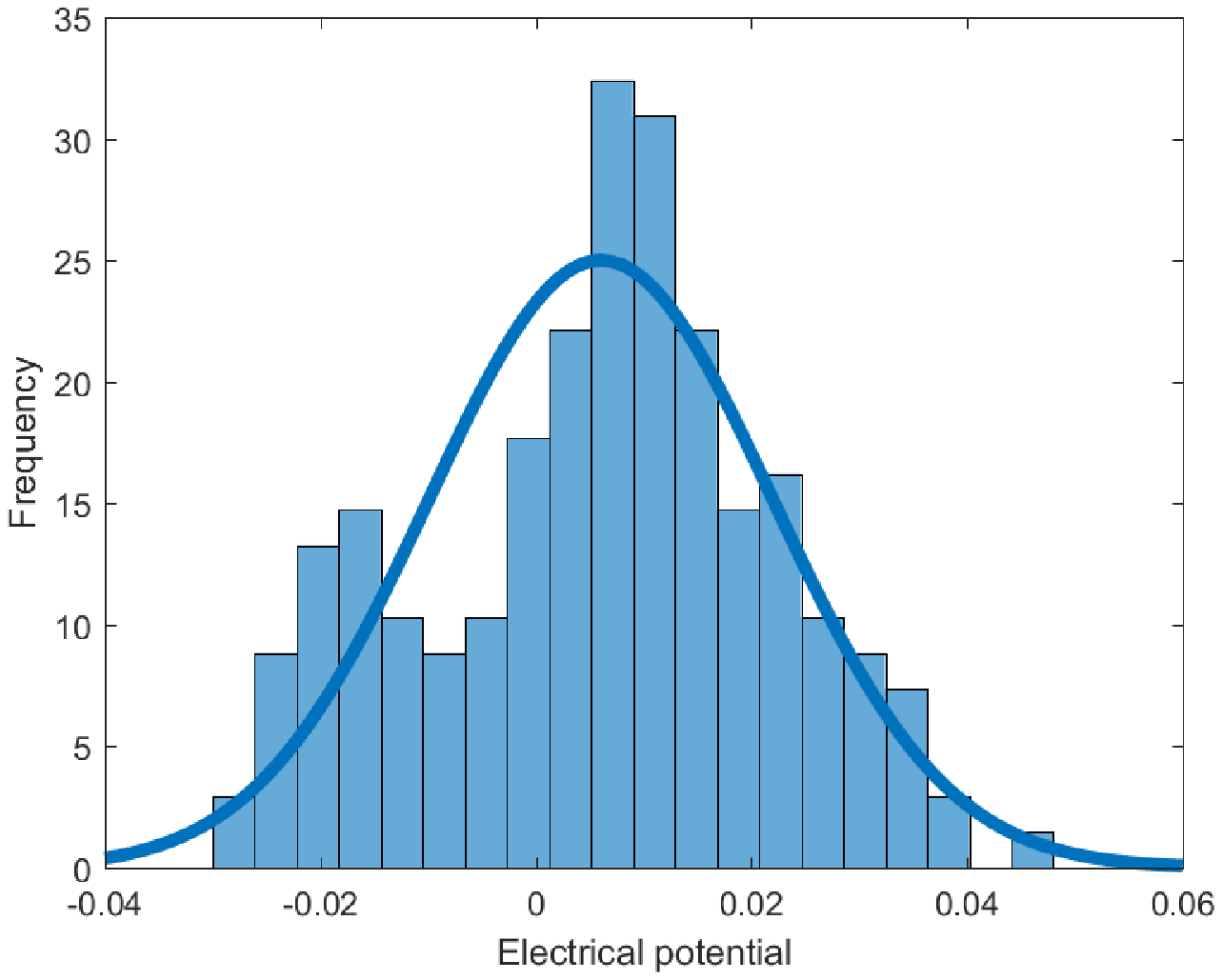}
    \caption{Histogram and fitted normal distribution of appearances of electrical potential of the BZ LM for all four experiments.}
    \label{fig:dist}
\end{figure}

\textcolor{black}{The electrodes are penetrating through the BZ LM and the plastic container onboard of the robot. Consequently, the BZ LM is not able to move freely in the plastic container. Given that the oscillation period of the potential is similar in experiments without movement (Fig.~\ref{fig:exp}) and with movement (Figs.~\ref{fig:aexp} to~\ref{fig:a4exp}), the vibrations from the robot seem not to be enough to characterize the LM as a well-mixed system. As a result, the BZ LM can be considered as a distributed-parameter system with local concentration gradients.}

\textcolor{black}{All the electrical potential values saved on the on-board SD card of the Arduino system were congregated and investigated. The resulted data set was used to produce the histogram presented in Fig.~\ref{fig:dist}. Moreover, a fitted normal distribution of the appearances of each batch of electrical potential was plotted in Fig.~\ref{fig:dist}. The mean value is 0.006 and the standard deviation 0.0159. Thus, the definition of assigning left or right turns with values around zero (which is close to the mean of 0.006) provides an almost evenly distributed motion towards both sides.}

\section{Discussions}
\label{sec3}

This work demonstrates that the BZ reaction can be directly incorporated into the electronic circuitry of a controller for a robot. Limitations imposed by earlier prototypes of liquid phase controllers, where robots were restricted to forward speeds of ca.\ 1cm/s and rotation speeds of ca.\ 1 degree/s~\cite{adamatzky2004experimental}, were alleviated. The additional benefits of the BZ LM system were that no optical interfaces were required to monitor the BZ LM controller and the geometries of the oxidation wave-fronts no longer need to be analysed. Hardware and software used in previous versions of the robot~\cite{adamatzky2004experimental,yokoi2004excitable} (a light placed underneath the reaction contained within a Petri dish, a serial connection to a PC and image processing algorithms) are not necessary as the BZ LMs are electrically connected to the micro-controller that delivers the trajectory of the robot. This reduction in the complexity of the controller system shows progress towards future unconventional and soft robotics.

%We demonstrated that BZ reaction can be directly incorporated into electronic circuit of a robot controller. Thus we alleviated limitations imposed by earlier prototypes of liquid phase controllers where a robot was restricted to forward speed (1~cm/s) and rotation speed (1~degree/s)~\cite{adamatzky2004experimental}. The additional benefits are that no optical interfaces are required to monitor the BZ LM controller and the geometries of the oxidation wave-fronts no longer needs to be analysed. Moreover, due to the fact that BZ LMs are electrically connected to the micro-controller that delivers the robot's trajectory, the additional hardware and software (light placed underneath the Petri dish, the serial connection to a PC and image processing algorithms) from the previous versions of the robot \cite{adamatzky2004experimental,yokoi2004excitable} are redundant.  This is a step further towards unconventional or soft robotics.

By encapsulating the BZ solution droplets in hydrophobic powder to form LMs, made the controllers re-configurable. In principle, it would be possible to mount as many BZ marbles as desired on-board of a robot and allow the LM ensembles to process information about the local environment and potentially make decisions based on the fusions of many stimuli. The properties of LMs can be tailored for a variety of applications by altering the encapsulated liquid and / or the powder coating~\cite{Bormashenko2017, Bormashenko2012, Fujii2016, McHale2015, Ooi2015, rychecky2017spheroid}. This means LMs can be prepared to enable them to be manipulated using electrical and magnetic fields, in addition to mechanical manipulation. Thus, robotic BZ LM controllers can be reconfigured on-flight, during the robot is in motion.

%By encapsulating BZ in the hydrophobic powder we are making controllers re-configurable. In principle, it possible to mount as many BZ marbles as desired on-board of a robot and allow the marble ensembles to process information about environment and, potentially, to make decisions based on fusion of many stimuli. Moreover, properties of LMs can be tailored for a variety applications by altering the encapsulated liquid or the powder coating~\cite{Bormashenko2017, Bormashenko2012, Fujii2016, McHale2015, Ooi2015, rychecky2017spheroid}. Therefore the LMs can be manipulated using electric and magnetic fields, in addition to mechanical manipulation. Thus, robotic BZ LMs controllers can be reconfigured on flight, during the robot motion. 

\textcolor{black}{It is noteworthy that the implementation proposed here is not an ideal and ready-to-use solution. This is an initial study towards the control of robots with chemical reaction-diffusion systems through an electrical connection. It is a contribution in bringing important improvement in prototyping wet robotics bearing complex dynamics. The exact behavior of the chemical system is difficult to predict and to manipulate as noticed in the results from the experiments. Consequently, the reproduction of the results and the perfect manipulation of the potential’s oscillation were not extensively analysed in the context of this study but remain as aspects of future work.}

\textcolor{black}{Finally, the short time of the experiments with the BZ LM mounted on the robot are not because of reagents depletion, but in order to illustrate more efficiently the reaction of the marble to laser illumination.}

\section*{Acknowledgments}
This research was supported by the EPSRC with grant EP/P016677/1 ``Computing with Liquid Marbles''.

% \section*{Author contributions}

% MAT, CF and AA conceived the idea, designed and conducted the experiments and wrote the paper. 

% \section*{Competing interests}

% The authors have declared that no competing interests exist.

\section*{Supporting information}

Videos and snapshots of experiments can be found at \cite{BZRobotVideoZenodo}.

\nolinenumbers

% Either type in your references using
% \begin{thebibliography}{}
% \bibitem{}
% Text
% \end{thebibliography}
%
% or
%
% Compile your BiBTeX database using our plos2015.bst
% style file and paste the contents of your .bbl file
% here. See http://journals.plos.org/plosone/s/latex for 
% step-by-step instructions.
% 

\end{document}